\documentclass[journal]{IEEEtran}
\usepackage{xcolor}
\usepackage{times}
\usepackage{soul}
\usepackage{url}
\usepackage[hidelinks]{hyperref}
\usepackage[utf8]{inputenc}

\usepackage[small]{caption}
\usepackage{graphicx}
\usepackage{amsmath}
\usepackage{booktabs}
\usepackage{algorithm}
\usepackage{algorithmic}
\usepackage{times}
\usepackage{epsfig}
\usepackage{graphicx}
\usepackage{amsmath}
\usepackage{amssymb}
\usepackage{times}
\usepackage{helvet}
\usepackage{xcolor}
\usepackage{courier}
\usepackage{amsmath}
\usepackage{amssymb}
\usepackage{graphicx}
\usepackage{subfigure}
\usepackage{longtable}
\usepackage{rotating}
\usepackage{multirow}
\usepackage{balance}
\usepackage{url}
\usepackage{makecell}
\usepackage{balance}
\makeatother

\usepackage{amssymb}

\ifCLASSINFOpdf
\else
\fi

\begin{document}

\title{Discriminative Cross-Domain Feature Learning for Partial Domain Adaptation}

\author{Taotao~Jing,~Ming Shao~\IEEEmembership{Member, IEEE}, ~Zhengming~Ding~\IEEEmembership{Member, IEEE},% <-this % stops a space

%\IEEEcompsocitemizethanks
%{\IEEEcompsocthanksitem

\IEEEcompsocitemizethanks{\IEEEcompsocthanksitem T. Jing, is with the Department of Electrical and Computer Engineering, Indiana University-Purdue University Indianapolis, Indianapolis, IN 46202, USA (e-mail: jingt@iu.edu)
\IEEEcompsocthanksitem M. Shao is with the Department of Computer and Information Science, University of Massachusetts Dartmouth, MA, 02747, USA.
(e-mail: mshao@umassd.edu).
\IEEEcompsocthanksitem Z. Ding is with the Department of Computer, Information and Technology, Indiana University-Purdue University Indianapolis, Indianapolis, IN 46202, USA. (e-mail: zd2@iu.edu).
}
}

%\markboth{JOURNAL OF \LaTeX CLASS FILES, VOL. **, NO. *, *** 2020}

\maketitle

\begin{abstract}
Partial domain adaptation aims to adapt knowledge from a larger and more diverse source domain to a smaller target domain with less number of classes, which has attracted appealing attention. Recent practice on domain adaptation manages to extract effective features by incorporating the pseudo labels for the target domain to better fight off the cross-domain distribution divergences. However, it is essential to align target data with only a small set of source data. In this paper, we develop a novel Discriminative Cross-Domain Feature Learning (DCDF) framework to iteratively optimize target labels with a cross-domain graph in a weighted scheme. Specifically, a weighted cross-domain center loss and weighted cross-domain graph propagation are proposed to couple unlabeled target data to related source samples for discriminative cross-domain feature learning, where irrelevant source centers will be ignored, to alleviate the marginal and conditional disparities simultaneously. Experimental evaluations on several popular benchmarks demonstrate the effectiveness of our proposed approach on facilitating the recognition for the unlabeled target domain, through comparing it to the state-of-the-art partial domain adaptation approaches.
\end{abstract}


% \fi
\IEEEpeerreviewmaketitle

\section{Introduction}

\begin{figure}[t]
\centering
\includegraphics[scale=0.275]{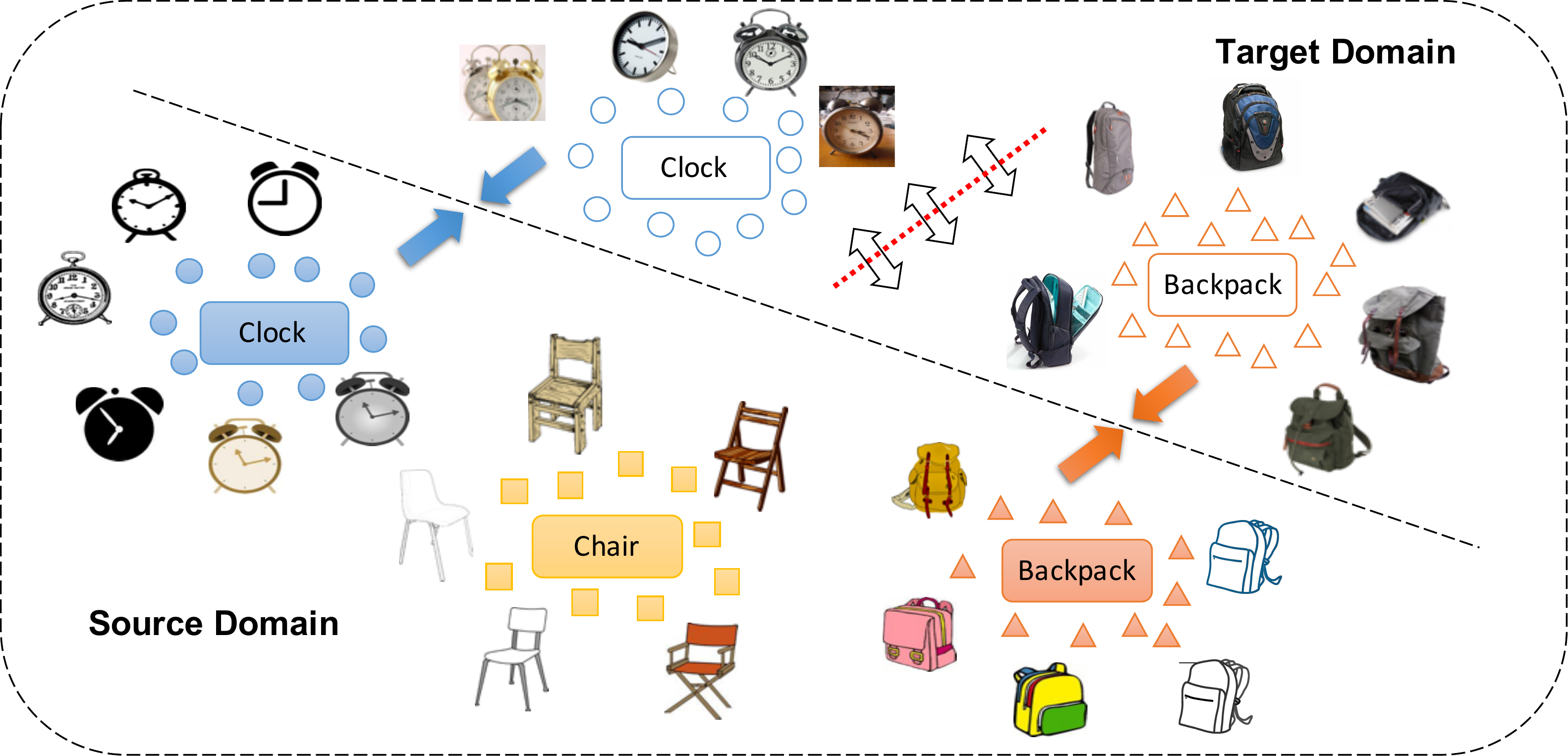}
    \caption{Illustration of partial domain adaptation, where external source domain covers more classes than unlabeled target domain. The goal is to learn more discriminative information by mitigating domain difference and removing irrelevant knowledge.}
    \label{fig:figure1} \vspace{-5mm}
\end{figure}

\IEEEPARstart{D}{omain} adaptation has cast a light to recognize the unlabeled target data with the help of knowledge transferred from an external well-established, but differently distributed source domain data \cite{long2013transfer,baktashmotlagh2013unsupervised,hou2016unsupervised,tsai2016domain,gholami2017punda,yan2017mind, cheng2014semi, niu2016exemplar,zhang2019bridging,8370105,8946732,8902166,7428839}. The mechanism of domain adaptation is to reveal the common latent factors between the source and target domains and explore them to reduce both the marginal and conditional mismatch in terms of the feature space across domains simultaneously. Traditional domain adaptation assumes the external source domain has the same category information with the target domain. However, as shown in Fig. \ref{fig:figure1}, real-world well-labeled source domains contain more categories than what we are targeting at in some cases, which results in a partial domain adaptation problem. Therefore, how to adapt the useful information from a large-scale source domain while removing the irrelevant knowledge becomes a key issue in partial domain adaptation problems. 

% \begin{figure}
%     \centering
%     \includegraphics[width=24em]{DCDF.png}
%     \vspace{-2mm}\caption{Illustration of our proposed method, which projects source and target samples to a domain-inviriant feature space, where unlabeled target samples will be aligned only to shared classes between source and target label space, while outlier classes will be ignored.}
%     \label{fig:DCDF}
% \end{figure}

Recent research efforts on deep neural networks (DANN) discover that deep structure learning can capture domain-invariant features for knowledge transfer with promising performance on existing cross-domain benchmarks \cite{krizhevsky2012imagenet,simonyan2014very,yan2016image,li2018heterogeneous,li2016prediction, ben2007analysis,li2019joint,zhang2019domain,8964455,8740907,8946732}. Specifically, deep structure learning manages to unfold exploratory factors of variations within the data, and cluster representations layer by layer according to their similarity \cite{bengio2013representation}. However,  with the domain discrepancy enlarged, feature transferability drops significantly in the top task-specific layers \cite{long2015learning,tzeng2015simultaneous}. In other words, the features extracted from the top task-specific layers highly depend on the source data distribution, which is not valid for unseen differently distributed target domains.

Most recently, partial domain adaptation becomes an immediate area of research focus, which assumes a large-scale source domain is diverse enough to subsume all classes in a small-scale target domain of interest. Furthermore, the target domain data are not only unlabeled, while we further have no idea about the size of the target domain label space nor the corresponding categories. Intuitively, merely aligning the whole source and target domains is not a good enough solution to address partial domain adaptation problems, since mixing the irrelevant source sub-classes with the target data together may result in the degradation of the target classification performance. Thus, filtering out the irrelevant source classes and enhancing the most similar source classes effects with the target domain is crucial for effective knowledge transfer. To achieve this purpose, \cite{SAN,PADA} propose to maximally align both domains' distributions in the shared label space while diminishing the negative impact of irrelevant source classes. \cite{IWAN, PADA} identify the importance weight of each source data with the help of a different adversarial domain classifier automatically. Unfortunately, the adversarial network based models would have a lot of weights and parameters need to train and optimize. \cite{li2020deep} proposes a special framework equipped with a residual block along with the task-specific feature layer to promote the features representation capability for cross-domain adaptation and effectively weakens the negative transfer caused by the irrelevant classes.

% However, only recognizing shared and outlier classes, while learning classifier over the whole source data would still mislead the classification on the target domain. {\color{red} \cite{li2020deep} DRCN proposes ... }
% % \cite{ETN} propose different frameworks to simultaneously quantify source samples transfer-ability and down-weighting the classifier
% Unfortunately, the adversarial network based models would have a lot of weights and parameters need to train and optimize.

In this paper, we propose a novel partial domain adaptation model via a newly-designed weighted cross-domain center loss and cross-domain graph propagation in an EM-like optimization strategy. The key idea of our model is to seek a domain-invariant feature space, where unlabeled target data are well-aligned with relevant source data while outlier source classes' influence will be removed. To sum up, our contributions are listed in two folds:
\begin{itemize}
    \item We propose a cross-domain center loss to seek a domain-invariant feature space, where the unlabeled target samples tend to be coupled with different source class centers in a probabilistic reconstruction format. Through optimizing the reconstruction coefficients, relevant source centers would have higher reconstruction coefficients, while irrelevant source centers would have smaller ones. 
    \item We propose a weighted cross-domain graph to propagate the relevant source labels to the unlabeled target samples. The cross-domain graph could help capture the intrinsic structure within the source and target to emphasize more on relevant source classes in label propagation. Thus, the predicted target labels could be further fed to the cross-domain center loss to optimize feature learning.
\end{itemize}

The remaining sections of this paper are organized as follows. In Section II, we provide a brief review of the related works and highlight the differences. We present our novel discriminative cross-domain feature learning framework in Section III, as well as the solution and complexity analysis of our method. Experimental analyses are provided in Section IV, followed by the conclusion in Section V.


%\IEEEpubidadjcol

\section{Related Work}

%% Domain adaptation review
\subsection{Domain Adaptation}
To manage the distribution difference between domains, prior efforts on domain adaptation usually attempt to alleviate the domain discrepancy through instance reweighting and domain-invariant feature learning. Instance reweighting based methods try to reweight each source domain sample to align the source and target domain distribution. Nevertheless, these methods fail when the source and target data drawn from the different conditional distributions. On the other hand, feature learning-based methods are encouraged to derive domain-invariant features or latent subspaces to match the distribution disparity across domains. Among them, subspace-based efforts have achieved promising results by exploring a domain-invariant low-dimensional feature space to align the two different domains.

Deep neural networks have achieved remarkable advances in classic classification tasks. However, for different distribution data, the domain discrepancy is enlarged at the top layers, which fails the generally trained classifier, and to address this issue, deep domain adaptation methods aim to explore a deep end-to-end architecture to mitigate the domain shift jointly\cite{ganin2015unsupervised,long2015learning,tzeng2015simultaneous,sun2016deep,duan2012domain, li2016prediction}. Generally, MMD or revised MMD loss \cite{long2015learning,yan2017learning}, and adversarial loss  \cite{tzeng2015simultaneous} are popular strategies to eliminate the domain shift with deep structures. However, prior deep domain adaptation algorithms ignore the conditional distribution divergence across domains, only seek to couple the source and target domains as a whole. Taking the conditional distribution into account and incorporating class-wise alignment for activate feature learning is appropriate and straightforward.

%% Partial domain adaptation review
\subsection{Partial Domain Adaptation}
With the development of big data techniques and more large-scale datasets available, it is realistic to require us to transfer partial relevant knowledge from the source to the small-scale unlabeled target domain dataset. Previous domain adaptation approaches assuming the source and target domain have identical label space, which is vulnerable to negative transfer in the partial transfer problems. \cite{SAN} proposes a Selective Adversarial Network (SAN) to address the partial domain adaptation problems through reweighting each sample and maximally align the data distributions across domains in the shared label space, which benefits transferring relevant data and eliminating drawbacks of irrelevant data simultaneously. Partial Adversarial Domain Adaptation (PADA) alleviates the negative transfer by down-weighting the data of outlier source classes \cite{PADA}. Importance Weighted Adversarial Networks \cite{IWAN} presents an adversarial nets-based framework to quantify the importance of each source sample and recognize those potentially from the outlier classes, then reduce the domain shift of the shared classes across domains. With the help of adversarial networks and min-max optimization strategies, these methods achieve significantly better performance than classical domain adaptation models. The latest work Deep Residual Correction Network (DRCN) \cite{li2020deep} implements residual block to boost the feature representation capability and designs a weighted class-wise domain alignment loss to match to cross-domain shared classes feature distributions. 

% However, recognizing source domain outlier classes from shared classes, while still training the classifier on the whole source data could distract the performance. 
% {\color{red} \cite{li2020deep} DRCN ... }
% To this end, Example Transfer Network \cite{ETN} proposes a new framework to quantify the source examples transferability and down-weighting irrelevant source examples from outlier classes on the source classifier.

%% Our work

Differently, we equip the subspace learning technique to align both marginal and conditional distribution disparity across the relevant source and target domain samples. Meanwhile,  a cross-domain graph built on the shared space can capture the intrinsic structure of the data distribution and better transfer the label information. Specifically, label propagation \cite{LabelPropagation} would be iteratively optimized with the domain-invariant feature learning framework to refine the class-wise adaption term. Exploring the contribution of the soft labels and their probability is not only needed but also effective.  This is the most significant difference compared to existing works and is our main contribution.

\section{The Proposed Method}

In this section, we first list our novel discriminative cross-domain feature learning framework. Then, we provide an efficient solution via EM optimization, as well as complexity analysis. Table \ref{notation} shows the frequently used notations.

\subsection{Preliminaries}
Given labeled source domain data $\mathcal{D}_s = \{ \mathbf{X}_s,\mathbf{Y}_s\} = \{(\mathbf{x}_s^1,\mathbf{y}_s^1), \cdots , (\mathbf{x}_s^{n_s},\mathbf{y}_s^{n_s}) \}$ where $\mathbf{x}_s^i \in \mathbb{R} ^{d}$ is a $d$-dimension source domain sample and $\mathbf{y}_s^i \in \{0,1\}^{C_s}$ is the associated label, $C_s = |\mathcal{C}_s|$ is the number of classes of source domain label space $\mathcal{C}_s$. $\mathcal{D}_t = {\mathbf{X}_t} = \{\mathbf{x}_t^1, \cdots  , \mathbf{x}_t^{n_t}\}$ is unlabeled target domain features with $\mathbf{x}_t^i \in \mathbb{R}^d$ is the $d$-dimension target domain feature without label. In classical domain adaptation tasks, source and target domains have different features distribution $P_s(\mathbf{x}_s) \neq P_t(\mathbf{x}_t)$, while identical label space $\mathcal{C}_s = \mathcal{C}_t$, $\mathcal{C}_t$ is the target domain label space. In this paper, we focus on partial domain adaptation problem when source domain label space subsumes target domain \cite{SAN,PADA}, i.e., $\mathcal{C}_s \supset \mathcal{C}_t$. Standard domain adaptation methods suffer from the negative transfer caused by outlier classes from the source domain. 

\begin{table}[t]
\begin{center}\caption{Notations and Descriptions.}
\label{notation}
\renewcommand{\arraystretch}{1.3}\vspace{-2mm}
\begin{tabular}{cl}
  \Xhline{1pt}
  Notation & Description\\
  \hline
    $\mathcal{D}_s, \mathcal{D}_t$ & source / target domain \\
    
    $\mathcal{C}_s, \mathcal{C}_t$ & source / target domain label space \\
    
    $\mathbf{X}_s, \mathbf{X}_t$ & source / target input matrix \\
    
    $\mathbf{Y}_s$  & source domain label distribution\\
    
    $\mathbf{P}_t, \hat{\mathbf{Y}}_t$ & predicted target  soft / hard label distribution\\
    
    $\mathbf{x}_s^i, \mathbf{x}_t^j$ & source / target domain instance \\

    $n_s, n_t$ & source / target samples number \\

    $d , k$ & original / embedding feature dimension \\

    {$\alpha_p, \alpha_c$, $\lambda$ } & balance factor of loss items\\
    
    $ G $ & weighted cross-domain graph \\
    
    $ \mathbf{W}$ & corresponding weights of G\\
    % $w_{ij}$ & similarity between node $i$ and $j$ \\
    
    $\mathbf{\bar{p}}_t$ & predicted class level weights\\

    $\omega_i$ & importance of source sample $\mathbf{x}_i$\\

    % $X$ & input data matrix \\

     $\mathbf{A}$ & Projection matrix \\

    % $Z$ & embedding matrix \\

    % $H$ & centering matrix \\

    % $M_c$ & MMD matrices, $c \in \left \{ 0, ... , C \right \}$\\

  \Xhline{1pt}
\end{tabular}
\end{center}\vspace{-6mm}
\end{table}

\subsection{Motivation}

Partial domain adaptation \cite{SAN,PADA,IWAN} assumes labeled source and unlabeled target domain have inconsistent label space, which makes it impractical to apply the classifier obtained from source data directly to target data. Moreover, existing domain adaptation methods seek to minimize the marginal and conditional distribution between source and target domain, while matching the target samples to the whole source label space \cite{long2013transfer}. However, data from source domain outlier classes, which are not shared with the target domain, would cause negative transfer during adaptation. So addressing target data to shared source label space is crucial to managing partial domain adaptation tasks.

To solve these difficulties, we propose a discriminative cross-domain feature learning framework, which seeks a latent common feature space across the source and target domains through a weight refined cross-domain center loss. In the latent common feature space, source and target domain have similar distributions and are able to have similar classifier weights on shared categories. To better propagate the labels from source to target in the newly-learned space, we propose a weighted cross-domain graph to assign a probabilistic label for each target sample in source domain label space $\mathcal{C}_s$. The graph label propagation would assist intrinsic structure preserving across two domains to reduce the influence of outlier source categories. Furthermore, predicted probabilistic labels will refine the weighted cross-domain center loss iteratively. Our goal is to learn a domain-shared projection $\mathbf{A} \in \mathbb{R}^{d\times{k}}$, which transforms source and target data into a domain-invariant space by jointly preserving the discriminative information and detecting irrelevant source classes.

\subsection{Discriminative Cross-Domain Alignment}

\noindent{\textbf{Selective Domain-Wise Adaptation}}: Domain adaptation methods always seek to minimize source and target domain marginal distribution distance and conditional distribution distance jointly. We seek a linear transformation $\mathbf{A}$ to extract domain-invariant features across two domains. Empirical Maximum Mean Discrepancy (MMD) is widely used to measure the distribution difference across two domains, which aims to alleviate marginal distribution divergence. Here we introduce the transferable source examples to define the weighted domain-wise adaptation as:

\renewcommand{\arraystretch}{1.3}
\begin{equation}
\label{Lm}
\begin{aligned}
    L_{m}
    &= \left \| \frac{1}{\sum\limits_{i=1}^{n_{s}}\omega_i} \sum\limits_{i=1}^{n_{s}} \mathbf{A}^{\top} \omega_i\mathbf{x}_{s}^{i}  -  \frac{1}{n_{t}} \sum\limits_{j=1}^{n_{t}}  \mathbf{A}^{\top} \mathbf{x}_{t}^{j}  \right \|_{2}^{2}\\
    &= \mathrm{tr}(\mathbf{A}^{\top}\mathbf{XM_0X}^{\top}\mathbf{A}),
\end{aligned}
\end{equation}
where $\omega_i$ denotes the importance of source sample $\mathbf{x}_s^i$ in knowledge transfer and we would provide the calculation in the following part. $\mathbf{X}$ is the concatenation of $\mathbf{X}_s$ and $\mathbf{X}_t$, i.e., $\mathbf{X} = [\mathbf{X}_s, \mathbf{X}_t] \in \mathbb{R}^{d \times (n_s+n_t)}$ and $\mathbf{x}_i, \mathbf{x}_j$ are the \textit{i-th} and \textit{j-th} columns of $\mathbf{X}$. $n_s$,$n_t$ denote the number of samples in source and target domain respectively. $\mathrm{tr}(\cdot)$ is trace of the matrix, which is equal to the sum of main diagonal elements. $\mathbf{M_0}$ denotes the domain-wise alignment matrix with each element defined as:
\renewcommand{\arraystretch}{1.3}
\begin{equation}
\label{m0}
    (\mathbf{M_0})_{ij} = \left\{
    \begin{matrix}
    \frac{1}{\big(\sum\limits_{i=1}^{n_{s}}w_i\big)\big(\sum\limits_{i=1}^{n_{s}}w_i\big)}  ,&  \mathbf{x}_i, \mathbf{x}_j \in \mathcal{D}_s\\ 
    \frac{1}{n_t n_t}   ,&  \mathbf{x}_i, \mathbf{x}_j \in \mathcal{D}_t\\ 
    \frac{-1}{\big(\sum\limits_{i=1}^{n_{s}}w_i\big) n_t} ,&  \mathrm{otherwise}
    \end{matrix}
    \right.
\end{equation}

\vspace{1mm}\noindent{\textbf{Class-wise Adaptation via Weighted Cross-Domain Center Loss}}: Our previous selective MMD only considers the marginal distribution across source and target domains. Some existing domain adaptation works also explore pseudo labels to align conditional distribution \cite{long2013transfer}. However, they mostly assume complete domain adaptation to align every category across two domains, which would hurt the target learning by involving the outlier source categories in conditional distribution alignment. To address this issue, we design a novel weighted cross-domain center loss to minimize the conditional distribution disparity between relevant source categories and target data as:
\renewcommand{\arraystretch}{1.3}
\begin{equation}
\label{Lp}
\begin{aligned}
L_{p} 
&=  \sum\limits_{i=1}^{n_t} \sum\limits_{c=1}^{C_s} \left \| \mathbf{A}^{\top} \mathbf{x}_{t}^{i} -  \mathbf{A}^{\top} \, \mathbf{\mu}_s^c \, p_{t}^{i(c)}  \right \|  _{2}^{2} \\
&= \left \| \mathbf{A}^{\top} ( \mathbf{X}_{t} - \mathbf{X}_{s} \mathbf{Y}_s(\mathbf{Y}_s^\top \mathbf{Y}_s)^{-1}\mathbf{P}_{t}) \right \| _{F}^{2} \\
&= \mathrm{tr}(\mathbf{A}^{\top} \mathbf{X} \mathbf{M}_p \mathbf{X}^\top \mathbf{A}),
\end{aligned}
\end{equation}
where $ \mathbf{\mu}_s^c$ denotes the \textit{c-th} class center of source domain, i.e., $ \mathbf{\mu}_s^c =\frac{1}{n_{s}^{(c)}} \sum_{ \mathbf{x}_{j} \in \mathcal{D}_{s}^{(c)} } \mathbf{x}_{j}$, and $\mathcal{D}_s^{(c)}$ is source domain belonging to the \textit{c-th} category. $n_s^{(c)}$ is the total number of source domain samples in the specific \textit{c-th} category $\mathcal{D}_s^{(c)}$. $\mathbf{p}_t^i = [p_t^{i(1)}, ..., p_t^{i(C_s)}]^\top \in \mathbb{R}^{C_s}$ is the probabilistic label of target domain sample $\mathbf{x}_t^i$, and $p_t^{i(c)}$ denotes the probability that $\mathbf{x}_t^i$ belongs to the \textit{c-th} category. $\mathbf{P}_t = \{ \mathbf{p}_t^1, ..., \mathbf{p}_t^{n_t} \} \in \mathbb{R}^{C_s \times n_t}$ is the collection of target domain samples probabilistic soft labels. Denoting $\mathbf{Y}_{st} = \mathbf{Y}_s(\mathbf{Y}_s^\top \mathbf{Y}_s)^{-1}\mathbf{P}_{t}$, then $\mathbf{M}_p \in \mathbb{R}^{(n_s+n_t) \times (n_s+n_t)}$ can be defined as:
\begin{equation}
\label{mp}
    \mathbf{M}_p = \left [
    \begin{matrix}
    \mathbf{Y}_{st} \mathbf{Y}_{st}^\top  ,& -\mathbf{Y}_{st}\\ 
    -\mathbf{Y}_{st}^\top   ,&  \mathbf{I}
    \end{matrix}
    \right ].
\end{equation}
where $\mathbf{I}$ is an identity matrix.

\vspace{1mm}\noindent{\textbf{Remark}}: Existing MMD based domain adaptation or partial domain adaptation methods seek to minimize the distribution divergence between source and target domains by incorporating pseudo-labels of target samples \cite{long2016unsupervised}. However, they only assign one hard pseudo-label to each target sample, while the inconsistency of source and target domain label space makes it easy to undermine the data structure within target domain, especially when the classifier performing poorly at the beginning of optimization. Hence we adopt the probabilistic label $p_t^{i(c)}$ for every target sample, which is iteratively predicted. $p_t^{i(c)}$ measures the similarity between target sample $\mathbf{x}_t^i$ and the source domain class center of category $c$. On the other word, $p_t^{i(c)}$ denotes the contribution of source domain class center $c$ to target sample $\mathbf{x}_t^i$ during domain alignment. When the predicted pseudo-label of target data is inaccurate, it will not destroy the domain adaptation thoroughly, due to the probabilistic soft label  $\mathbf{p}_t^i$ constraint.

\vspace{1mm}\noindent{\textbf{Discriminative Domain-Invariant Center Loss}}: To further minimize the difference of distribution from same class in difference domains,  we also accept the relaxed domain-irrelevant clustering-promoting term \cite{liang2018aggregating} to jointly pull the embedding class centers from same category closer, regardless of source or target domains. The same class-clustering encouraging loss term is:
\begin{equation}
    \label{Lc}
    \begin{aligned}
        L_{c} 
        &=  \sum_{c=1}^{C_s} \frac{1}{n_c} \sum_{x \in \mathcal{D}_c} \| A^\top ( x - \mu_c) \|_2^2 \\
        % &< \sum_{c=1}^{C_s} \frac{1}{n^c} \sum_{x \in D_c} \left (  \left \|x^c -\mu_s^c\right \|_2^2 + \left \| x^c - \mu_t^c \right \| \right )\\
        &= \left \| \mathbf{A}^{\top} ( \mathbf{X} - \mathbf{X} \mathbf{Y}(\mathbf{Y}^\top \mathbf{Y})^{-1}\mathbf{Y}^\top) \right \| _{F}^{2} \\
        & = \mathrm{tr} (\mathbf{A}^\top \mathbf{X} \mathbf{M}_c \mathbf{X}^\top \mathbf{A}),
    \end{aligned} 
\end{equation}
% \begin{equation}
%     \label{Lc}
%     \begin{aligned}
%         L_{c} 
%         &=  \sum_{c=1}^{C_s} \left ( \frac{1}{n^c} \sum_{x \in D_c} \| A^\top ( x - \mu_c) \|_2^2  + \left \| \mu_s^c - \mu_t^c  \right \|_2^2 \right ) \\
%         &< \sum_{c=1}^{C_s} \frac{1}{n^c} \sum_{x \in D_c} \left (  \left \|x^c -\mu_s^c\right \|_2^2 + \left \| x^c - \mu_t^c \right \| \right )\\
%         & = \mathrm{tr} (\mathbf{A}^\top \mathbf{X} \mathbf{M}_c \mathbf{X}^\top \mathbf{A}),
%     \end{aligned} 
% \end{equation}
where $\mathcal{D}_c$ is the group of samples belonging to class $c$ from both source and target domains. $\mu_c$ is the class center across two domains. Denote $\mathbf{Y}_c = \mathbf{Y}(\mathbf{Y}^\top \mathbf{Y})^{-1}\mathbf{Y}^\top$, where $\mathbf{Y = [Y_s; P_t]}$, then $\mathbf{M}_c$ is defined as:
\begin{equation}
    \label{Mc}
    \begin{aligned}
    \mathbf{M}_c = (\mathbf{I}-\mathbf{Y_c})(\mathbf{I}-\mathbf{Y_c})^\top.
        % \mathbf{M}_c = \left [
        % \begin{matrix}
        %     \mathbf{Y_c Y_c^\top} & -\mathbf{Y_c^\top} \\
        %     -\mathbf{Y_c}^\top & \mathbf{I}
        % \end{matrix}
        % \right ].
    \end{aligned}
\end{equation}

However, the source domain class centers belonging to outlier classes label space $\mathcal{C}_s \setminus \mathcal{C}_t$ would cause negative transfer problems. Moreover, due to the domain shift and features distribution mismatch, especially at the beginning of optimization, the predicted probabilistic labels $\mathbf{p}_t^i$ of target domain samples $\mathbf{x}_t^i$ could be inaccurate, which will also mislead the optimizing direction. In order to eliminate the influence from wrongly predicted samples, we explore the class level weights $\mathbf{\bar{p}}_t = \sum_{i=1}^{n_t}{\mathbf{p}_t^i}$, where $\mathbf{\bar{p}}_t \in \mathbb{R}^{C_s}$ is the predicted class level distribution of source domain classes in target domain.

Since $\mathbf{\bar{p}}_t$ is the overall estimation across all source categories to search the relevant ones, we consider extremely smaller values in $\mathbf{\bar{p}}_t$ denoting those outlier source categories. To further reduce the negative transfer in our designed cross-domain center loss, we propose a binary operation on each predicted probabilistic label $\mathbf{p}_t \leftarrow \mathbf{p}_t\odot \mathcal{B}(\mathbf{\bar{p}}_t, \delta)$, where $\mathcal{B}(,)$ is the binary operator on $\mathbf{\bar{p}}_t$ with threshold $\delta$. That is, if the element value in $\mathbf{\bar{p}}_t^i$ is greater than $\delta$, $\mathcal{B}(\mathbf{\bar{p}}_t^i, \delta) = 1$, otherwise 0. Such a binary operator would help  some wrong predictions assigned to outlier source categories, since the overall prediction would eliminate the wrongly predicted labels. Similarly, we could also apply the binary operation on Eq. \eqref{Lc} to remove outlier classes and get binary weighted loss term $L_c$. If we set $\delta=0$, we would still rely on the probabilistic labels to figure out the outlier source categories. When we enlarge $\delta$ to a smaller positive value, say $10^{-3}$, we can further incorporate the overall prediction to refine the probabilistic labels for an effective cross-domain alignment. We further define $\omega_i$ for each source sample simply based on binary class weight $\mathbf{\bar{p}}_t$, to denote the importance of each sample. That is, we span each element in $\mathbf{\bar{p}}_t$ to all the same-class samples in $\omega_i$. 

\vspace{1mm}\noindent{\textbf{Overall Objective Function}}: By integrating selective domain-wise adaptation, weighted class-wise alignment and the discriminative domain-invariant center loss, we obtain the objective function of domain-invariant feature learning as:
\renewcommand{\arraystretch}{1.3}
\begin{equation} 
\label{overall}
\begin{array}{c}
    \min\limits_{\mathbf{A, P}_t}~\mathrm{tr}(\mathbf{A^{\top}X M_{all} X^{\top}A})+ {\lambda \left \| \mathbf{A} \right\| }_{\mathrm{F}}^{2}\\
    \mathrm{s.t.}~~ \mathbf{A^{\top}XHX^{\top}A= \mathbf{I}},
    \end{array}
\end{equation}
% \begin{equation} 
% \label{overall}
% \begin{array}{c}
%     \min\limits_{\mathbf{A, P}_t}~~~\alpha~ \mathrm{tr}\Big(\mathbf{A}^\top(\mathbf{X}_{t} - \overline{\mathbf{X}_{s}}\mathbf{P}_{t}) (\mathbf{X}_{t} - \overline{\mathbf{X}}_{s} \mathbf{P}_{t})^{\top}\mathbf{A}\Big)\\ 
%     ~~~+ (1-\alpha)\mathrm{tr}(\mathbf{A^{\top}XMX^{\top}A})+ {\lambda \left \| \mathbf{A} \right\| }_{\mathrm{F}}^{2}\\
%     \mathrm{s.t.}~~ \mathbf{A^{\top}XHX^{\top}A= \mathrm{I}},\vspace{-1mm}
%     \end{array}
% \end{equation}
where  $\mathbf{M_{all}} = \mathbf{M}_0 + \alpha_p \mathbf{M}_p + \alpha_c \mathbf{M}_c$, and $\alpha_p, \alpha_c$ are balance factors. $\lambda$ is regularization parameter to $\|\mathbf{A}\|_F^2$, which is the Frobenius norm of $\mathbf{A}$, and $\mathbf{H} = \mathbf{I} - \frac{1}{n}\mathbf{1}$ denotes the centering matrix. The constraint seeks to maximize the embedded data variance \cite{long2013transfer}. For non-linear problems, we can apply kernel mapping $\mathbf{x} \mapsto \mathbf{\psi(x)}$ to construct kernel matrix $\mathbf{K} = \psi(\mathbf{X})^\top \psi (\mathbf{X}) \in \mathbb{R}^{n\times n}$, where $n = n_s + n_t$ and the kernel could be ``linear'' kernel \cite{long2013transfer}.

\subsection{Label Refinement via Cross-Domain Structural Knowledge}

The key challenges for partial domain adaptation are the target data is totally unlabeled and source domain contains outlier categories irrelevant to target data. Our previously designed cross-domain  center loss aims to mitigate the domain shift across two domains by assigning probabilistic labels to target samples. To further exploit the intrinsic structure across source and target domains while searching relevant source categories, we propose a weighted cross-domain graph $G$ to propagate the labels more likely from relevant source to target data. Actually, $G$ consists of four components $G_{ss}$, $G_{st}$, $G_{ts}$ and $G_{tt}$, which are within-source graph, source-target graph, target-source graph and within-target graph, respectively. 

When we propagate the label from source to target, we would only consider $G_{ts}$ and $G_{tt}$ \cite{LabelPropagation}. Suppose the weights for $G_{ts}$ and $G_{tt}$ are $\mathbf{W}_{ts}$ and $\mathbf{W}_{tt}$, we can predict target samples with the label information of source $\mathbf{Y}_s$ as
\begin{equation}\label{labelpropagation}
    \mathbf{P}_t = (\mathbf{I} - \mathbf{W}_{tt})^{-1} \mathbf{W}_{ts} \mathbf{Y}_{s},
\end{equation}
where we define the weights of two matrices $\mathbf{W}_{ts}$ and $\mathbf{W}_{tt}$ with $w_{ij} = \exp(-\frac{d_{ij}^{2}}{\sigma^{2}})$, which measures the similarity between node $i$ and $j$. Specifically, $d_{ij}$ is the distance between two samples under the learned domain-invariant space. The closer the nodes are, the edge and weight $w_{ij}$ are larger, which allows label propagating easier. To further eliminate the negative transfer from irrelevant source categories iteratively, we intuitively explore the overall prediction $\mathbf{\bar{p}}_{t}$ to reduce the impact of the outlier source categories in $\mathbf{W}_{ts}$. More specifically, $\mathbf{W}_{ts}^{i,j}$ links $i$-th target sample and $j$-th source sample from $c$-th class. Thus, $\mathbf{W}_{ts}^{i,j} \leftarrow \mathbf{W}_{ts}^{i,j} \mathbf{\bar{p}}_{t}^c$.

\begin{algorithm}[t]
    \caption{Proposed DCDF Framework}
    %\caption{Solution to the proposed model}
    \label{algorithm1}
  \begin{algorithmic}[1]
    \STATE \textbf{Input} Source and target feature matrices $\mathbf{X}_s$ and $\mathbf{X}_t$, source domain labels $\mathbf{Y}_s$ 
    \STATE \textbf{Initialization}: Construct $\mathbf{X}=[\mathbf{X}_s,\mathbf{X}_t]$ and initialize the weighted cross-domain graph $G$ on original features $\mathbf{X}$. Predict target domain data probabilistic labels $\mathbf{P}_t$ through by Eq.(\ref{labelpropagation}), and calculate class-wise binary weights $\mathbf{\bar{p}}_t$.  %, and initialize $\mathbf{M}_0, \mathbf{M}_p, \mathbf{M}_c $. 
    \REPEAT
    \STATE Solve the eigen-decomposition problem in Eq.(\ref{updateA}) and select the $k$ smallest eigenvectors to construct the adaptation matrix $\textbf{A}$
    \STATE Update target domain probabilistic labels $\mathbf{P}_t$, class-wise binary weights \textbf{$\mathbf{\bar{p}}_t$} by label propagation on new projected embedding features {$\mathbf{Z} = [\mathbf{A}^\top \mathbf{X}_s, \mathbf{A}^\top \mathbf{X}_t]$}
    \UNTIL{Convergence}
    \STATE \textbf{Output} Projection matrix $\mathbf{A}$, predicted labels $\mathbf{\hat{Y}}_t$ of target domain data 
    
  \end{algorithmic}
\end{algorithm}

\subsection{Optimization}

It is easy to check that $\mathbf{A}$ and $\mathbf{P}_t$ in Eq. \eqref{overall} cannot be jointly optimized. However, it is solvable over each of them in a leave-one-out manner. Specifically, we explore an EM-like optimization scheme to update the variables. For \textbf{E-step}, we fix $\mathbf{P}_t$ and optimize domain-invariant transformation $\mathbf{A}$; while for \textbf{M-step}, we update the probabilistic labels $\mathbf{P}_t$ with $\mathbf{A}$ fixed. Hence, we optimize two sub-problems iteratively.

% To optimize Eq.(\ref{overll}), we define $\Phi = diag(\phi_{1}, ... , \phi _{k}) \in \mathbb{R}^{k \times k}$ as the Lagrange multiplier, and derive the Lagrange function as
\vspace{1mm}\noindent{\textbf{E-step}:} With $\mathbf{P}_t$ fixed, we can optimize $\mathbf{A}$ with the following 
% \begin{equation}
% \label{lagrange_function}
%     \begin{aligned}
%     L = &tr(A^{T}((1-\alpha)XM_{0}X^{T}
%       + \alpha (X_{t} - \overline{X_{s}}P'_{t}) (X_{t} - \overline{X_{s}} P'_{t})^{T} \\ &+ \lambda I)A)
%       + tr((I - A^{T}XHX^{T}A)\Phi)
%     \end{aligned}
% \end{equation}
% Set $\frac{\partial L}{\partial A} = 0$, we obtain 
generalized eigen-decomposition problem:
\begin{equation}
\label{updateA}
    \Big(\mathbf{X}(\mathbf{M}_0 + \alpha_p \mathbf{M}_p + \alpha_c \mathbf{M}_c) \mathbf{X}^\top +\lambda \mathbf{I}\Big)\mathbf{A}
    = \mathbf{XHX^{\top}A\Phi},
\end{equation}
where we define the vectors $\mathbf{a}_i~(i \in [0, k$-$1])$ are obtained according to its minimum eigenvalues. Thus, we achieve $\mathbf{A} = [\mathbf{a}_0,\cdots,\mathbf{a}_{k-1}]$. $k$ is the embedding features dimension. 

% and the $k$ smallest eigenvectors of Eq.(\ref{eigendecomposition}) is the optimal adaptation matrix  $A \in \mathbb{R}^{D \times k}$.

% \textbf{Kernelization:} Considering kernel mapping $\psi(X) = [\psi(x_1), ... , \psi (x_n)]$ , and kernel matrix $K = \psi(X)^T\psi(X) \in \mathbb{R}^{n \times n}$ to non-linear problems, we can represent Eq.(\ref{eigendecomposition}) as:
% \begin{equation}
% \label{kernel_eigendecomposition}
%     \begin{aligned}
%     ((1-\alpha)KM_{0}K^{T}
%       + \alpha (K_{t} - \overline{K_{s}}P'_{t}) (K_{t} - \overline{K_{s}} P'_{t})^{T} + \lambda I)A\\
%       = KHK^{T}A\Phi
%       \end{aligned}
% \end{equation}

\vspace{1mm}\noindent{\textbf{M-step}:} The optimal adaptation matrix $\mathbf{A}$ will project source and target domain features to a new latent feature space, where we update the graph to optimize the predicted probabilistic labels $\mathbf{P}_t$ by Eq. \eqref{labelpropagation}. $\mathbf{P}_t$ and $\mathbf{A}$ will be optimized iteratively until converge.

By alternating the \textbf{E} and \textbf{M} steps detailed above, we optimize the problem iteratively until the objective function converges. It is noteworthy that the probabilistic labels of the unlabeled target samples are available with the valid domain-invariant projections. Then with the help of the label assignment strategy (Eq. \eqref{labelpropagation}), we can improve the projection discriminability (Eq. \eqref{updateA}) iteratively and refine the labeling quality and feature learning alternatively. We accept the label propagation method (Eq. \eqref{labelpropagation}) with $G$ built on the original features of the source and target domain to initialize $\mathbf{P}_t$. 

In summary, our proposed discriminative cross-domain feature learning framework is presented as Algorithm \ref{algorithm1}.

\begin{figure}[t]
    \centering
    \includegraphics[width=1\linewidth]{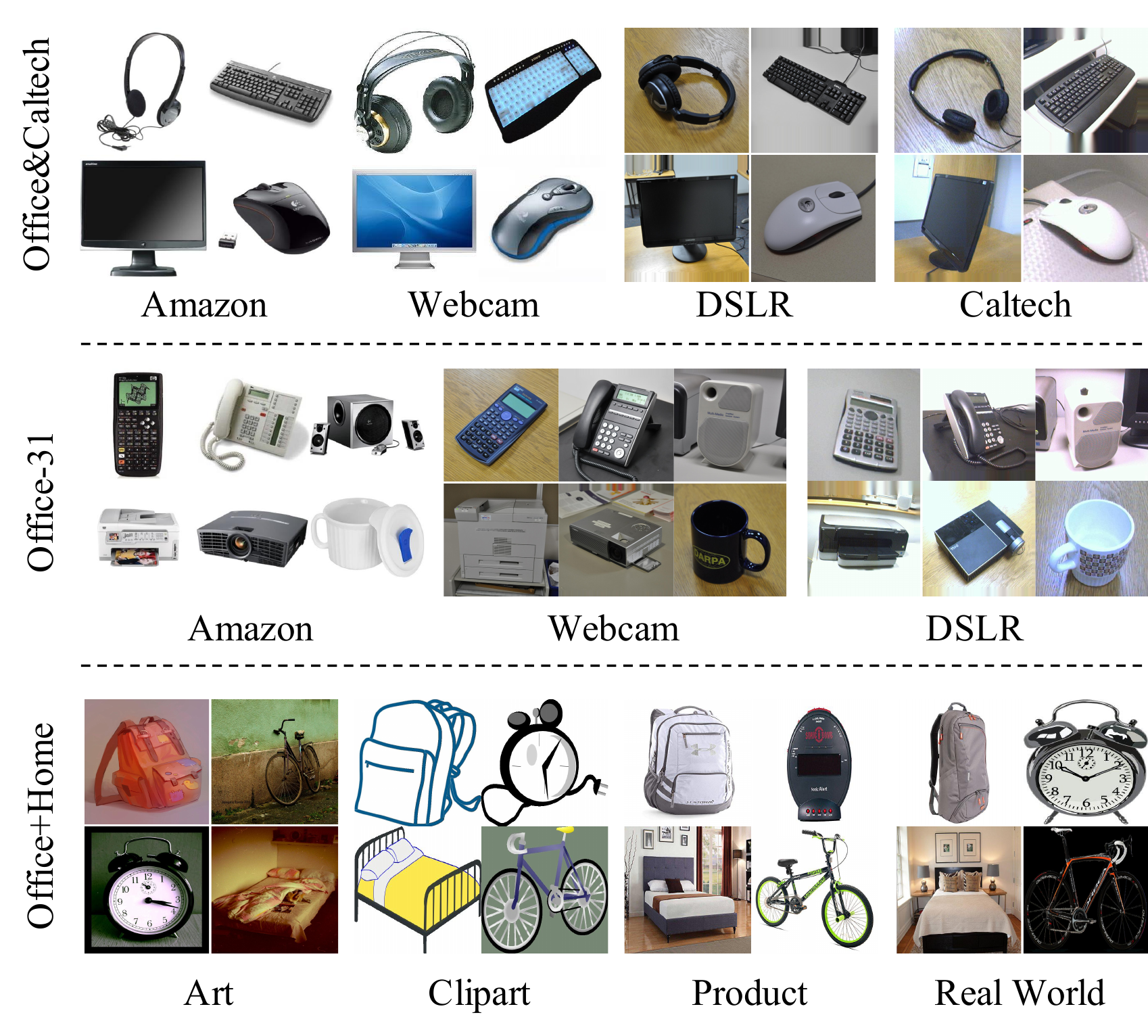}
    \caption{Example images of datasets Office\&Caltech, Office-31, and Office+Home.}
    \label{fig:datasets}
\end{figure}

\begin{figure}[t]
    \centering
    \includegraphics[width=1\linewidth]{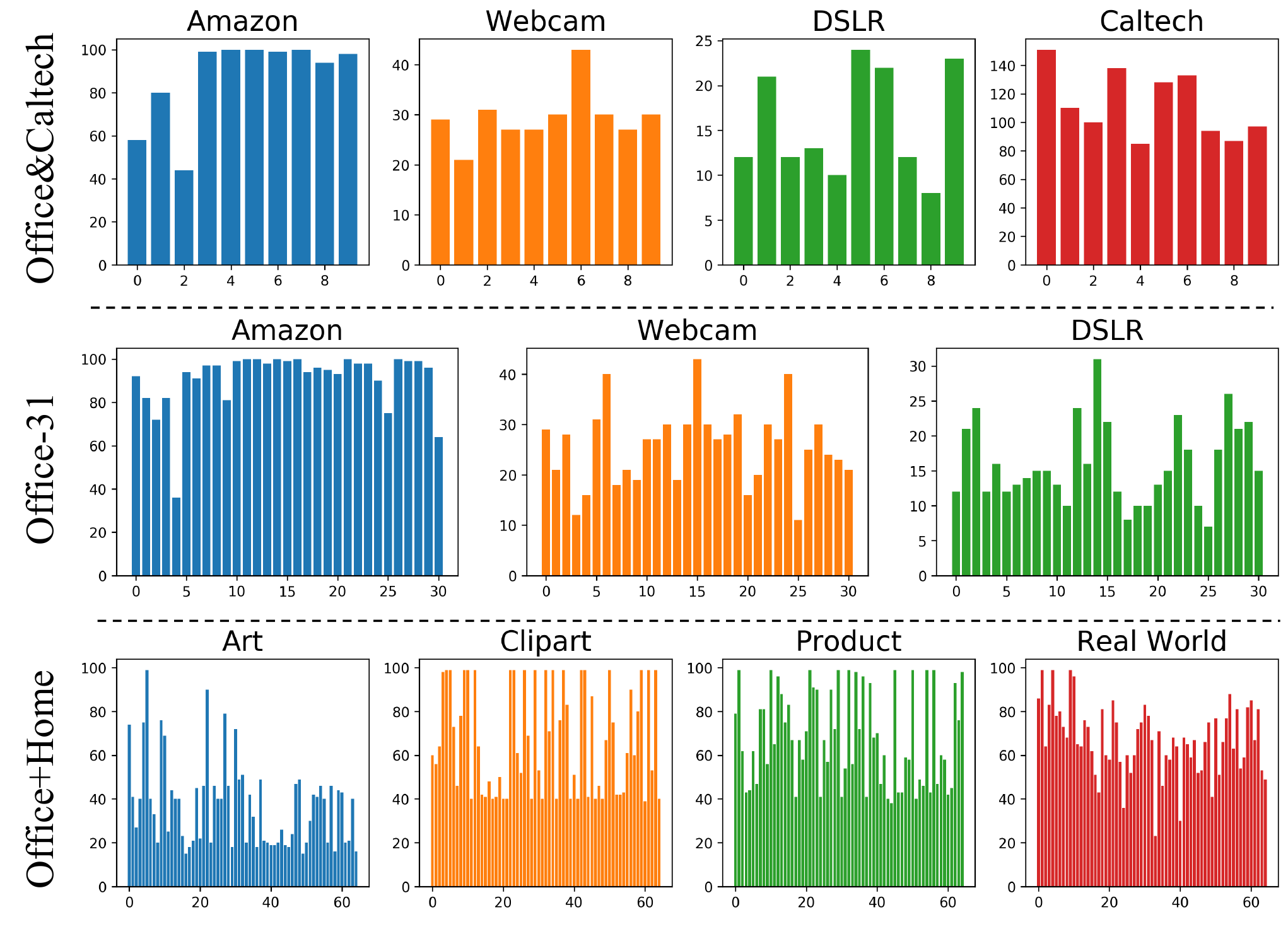}
    \caption{Number of samples belonging to each category from different domains of datasets Office\&Caltech, Office-31, and Office+Home.}
    \label{fig:datasets_dis} \vspace{-5mm}
\end{figure}

\begin{table*}

\centering
\caption{Comparisons of Recognition Rates ($\%$) of Partial Domain Adaptation on Office\&Caltech-10 Dataset (AlexNet)}
\label{OfficeCaltech10-5}
\setlength{\tabcolsep}{3pt} % Default value: 6pt
\renewcommand{\arraystretch}{1} % Default value: 1
%\begin{tabular}{|c|p{1cm}cc|ccc|ccc|ccc|c|}
\begin{tabular}{|c|p{1cm}p{1cm}p{1cm}|p{1cm}p{1cm}p{1cm}|p{1cm}p{1cm}p{1cm}|p{1cm}p{1cm}p{1cm}|c|}
\hline
     Method & A10 $\rightarrow$ W5 & A10 $\rightarrow$ D5  &  A10 $\rightarrow$ C5 &  W10 $\rightarrow$ A5  &  W10 $\rightarrow$ D5 &  W10 $\rightarrow$ C5  &  D10 $\rightarrow$ A5 &  D10 $\rightarrow$ W5  &  D10 $\rightarrow$ C5 &  C10 $\rightarrow$ A5  &  C10 $\rightarrow$ W5 &  C10 $\rightarrow$ D5  &  Avg  \\
\hline
% TCA & -- & -- & -- & -- & -- & -- & -- & -- & -- & -- & -- & -- & 87.29 \\ 
% GFK & -- & -- & -- & -- & -- & -- & -- & -- & -- & -- & -- & -- & 78.34 \\ 
% JDA & -- & -- & -- & -- & -- & -- & -- & -- & -- & -- & -- & -- & 86.74 \\ 
% TJM & -- & -- & -- & -- & -- & -- & -- & -- & -- & -- & -- & -- & 82.73\\ 
AlexNet \cite{krizhevsky2012imagenet} & \underline{76.30} & 85.29 & 85.27 & 87.37 & \textbf{100.0} & 74.14 & 89.51 & \underline{98.52} & 80.82 & 93.58 & 83.70 & 91.18 & 87.14 \\ 
RevGrad \cite{ganin2015unsupervised} & 65.93 & 80.88 & 77.57 & 80.30 & \underline{95.59} & 72.60 & 77.09 & 80.74 & 69.35 & 91.86 & 82.22 & 83.82 & 79.83 \\ 
RTN \cite{long2016unsupervised} & 69.63 & 70.59 & 80.99 & 74.73 & \textbf{100.0} & 59.08 & 70.02 & 91.11 & 59.08 & 91.86 & 93.33 & 80.88 & 78.44 \\ 
ADDA \cite{ADDA} & \textbf{87.41} & \underline{89.71} & 85.27 & 92.08 & \textbf{100.0} & 86.82 & 93.79 & \underline{98.52} & 89.90 & 93.15 & 94.07 & \underline{97.06} & 92.31 \\ 
IWAN \cite{IWAN}& \textbf{87.41} & 88.24 & \underline{89.90} & \underline{95.29} & \textbf{100.0} & \textbf{90.24} & \underline{94.43} &\underline{98.52} & \underline{91.61} & \underline{94.22} & \textbf{97.78} & \textbf{98.53} & \underline{93.84} \\ 
%PADA(Res50) & -- & -- & -- & -- & -- & -- & -- & -- & -- & -- & -- & -- & 00 \\ 

% Ours1 & \textbf{87.41} & 92.65 & 90.06 & 77.73 & \textbf{100} & 76.71 & 93.79 & 99.25 & 90.41 & \textbf{94.43} & 88.15 & 89.70 & 90.02\\

Ours & \textbf{87.41} & \textbf{94.12} & \textbf{92.12}& \textbf{95.50} &\textbf{100.0} & \underline{88.87} & \textbf{94.65} & \textbf{100.0} & \textbf{92.29} & \textbf{94.43} & \underline{94.81} & \underline{97.06} & \textbf{94.27}\\
\hline
\end{tabular}
\end{table*}
\begin{table*}

\centering
\caption{Comparisons of Recognition Rates ($\%$) of Partial Domain Adaptation on Office-31 Dataset (AlexNet)}
\label{Office31-10}
\setlength{\tabcolsep}{3pt} % Default value: 6pt
\renewcommand{\arraystretch}{1} % Default value: 1
\begin{tabular}{|c|cccccc|c|}
\hline
     Method & A31 $\rightarrow$ W10 & A31 $\rightarrow$ D10 &  W31 $\rightarrow$ A10  &  W31 $\rightarrow$ D10 &  D31 $\rightarrow$ A10 &  D31 $\rightarrow$ W10  &  Average  \\
\hline
AlexNet \cite{krizhevsky2012imagenet}]& 62.03 & 71.97 & 62.94 & 97.45 & 68.27 & 95.25 & 76.32 \\ 
DAN \cite{long2015learning}& 46.44 & 42.68 & 65.34 & 58.60 & 65.66 & 53.56 &55.38\\
RevGrad \cite{ganin2015unsupervised}& 56.95 & 57.32 & 63.15 & 89.17  & 57.62 & 75.59 & 66.64 \\
RTN \cite{long2016unsupervised}& 68.14 & 69.43 & 77.35 & 98.09 & 68.27 & 91.53 & 78.80 \\ 
ADDA \cite{ADDA}& 63.39 & 73.25 & 72.34 & 98.73 & 70.46 & 98.31 & 79.41 \\ 
SAN \cite{SAN}& \underline{80.02} & \underline{81.28} & \textbf{83.09} & \textbf{100.0}  & 80.58 & 98.64 & 87.27 \\ 
IWAN \cite{IWAN}& 76.27 & 78.98 & 81.73 & \textbf{100.0} & \underline{89.46} & \underline{98.98} & \underline{87.57} \\ 
%PADA(Res50) & 86.54 & 82.17 & 95.41 & 100 & 92.69 & 99.32 & 92.69 \\ 
% Ours1 & 85.08 & \textbf{87.26} & 81.94 & \textbf{100} & \textbf{92.90} & 98.98 & 91.03 \\  
Ours & \textbf{88.81} & \textbf{86.62} & \underline{82.00} & \textbf{100.0} & \textbf{91.23} &\textbf{99.66} & \textbf{91.39} \\
\hline
\end{tabular}
 \vspace{-3mm}
\end{table*}

\subsection{Complexity Analysis}

% In this section, we analyze our model complexity. 
There are two main time-consuming components: 1) $\mathbf{A}$ learning (\textbf{E-Step}); 2) $\mathbf{P}_t$ optimization (\textbf{M-Step}).

In detail, \textbf{E-Step} could cost $\mathcal{O}(d^3)$ for the generalized Eigen-decomposition of Eq. \eqref{updateA} for matrices with size of $\mathbb{R}^{d\times{d}}$, which could be reduced to $\mathcal{O}(d^{2.376})$ through the Coppersmith-Winograd method \cite{coppersmith1987matrix}. \textbf{M-Step} suffers from matrix multiplications. Generally, the multiplication for matrix with the size $n_t\times{n_t}$ could cost $\mathcal{O}(n_t^3)$. Assuming there are $l$ multiplication operations, \textbf{M-Step} would cost $\mathcal{O}(ln_t^3)$. Furthermore, we can speed up the operations of large matrices through a sparse matrix, and state-of-the-art divide-and-conquer approaches. Meanwhile, we could also store some intermediate computation results which could be reused in every stage.

\section{Experiments}

To illustrate the superiority of our model, we evaluate our proposed framework on several different partial domain adaptation tasks on three popular cross-domain benchmarks: \textbf{Office\&Caltech-10} \cite{gong2012geodesic}, \textbf{Office-31} \cite{Office31}, and \textbf{Office+Home} \cite{OfficeHome}. Sample images of each dataset from different domains are showed in Fig. \ref{fig:datasets}, and Fig. \ref{fig:datasets_dis} is the detailed class-wise data distribution of each dataset and domain, where the same domain different categories numbers of samples are sorted and displayed by the class labels alphabetically.

\noindent\textbf{Office\&Caltech-10} consists of 10 categories images from 4 domains: Amazon, Webcam, DSLR, and Caltech. Specifically, 10 shared classes with 3 domains from Office-31 \cite{Office31} dataset (Amazon, Webcam, DSLR) and 1 from Caltech-256 \cite{Caltech} (Caltech) constitute the Office\&Caltech-10 dataset. 12 different partial domain adaptation tasks are built by transferring from 10 classes source domain to 5 classes target domain \cite{IWAN}. %Select one domain as source domain - which is denoted as A10, W10, D10, and C10, the first 5 classes of another domain make up the target domain - which is denoted as A5, W5, D5, and C5. 

\noindent\textbf{Office-31} consists of 31 categories images from 3 domains: Amazon, Webcam, and DSLR.
%Fig. \ref{fig:datasets} shows samples from the four domains. 
Following the settings of \cite{SAN}, in each domain, we select the 10 shared classes between Office-31 and Office\&Caltech-10 as target and denote as A10, W10, and D10. Other domains with 31 classes constitute the source domain, which are denoted as A31, W31, and D31.% then we can come up with six partial domain adaptation tasks $A31 \rightarrow W10$, $A31 \rightarrow D10$, $W31 \rightarrow A10$, $W31 \rightarrow D10$, $D31 \rightarrow D10$, and $D31 \rightarrow W10$.

% For fair comparison, we follow the work \cite{long2013transfer} and \cite{pan2011domain} to evaluate our proposed model on 

% \begin{figure}[t]
%     \centering
%     \includegraphics[width=24em]{OfficeHome.png}
%     \vspace{-2mm}\caption{Example images from OfficeHome datasets.}\vspace{-1mm}
%     \label{fig:datasets2}
% \end{figure}

\noindent\textbf{Office+Home} is a larger domain adaptation benchmark, containing 65 different categories images from 4 domains: Art (Ar), Clipart (Cl), Product (Pr), and RealWorld (Rw).
%, and the sample images are showed in Fig. \ref{fig:datasets2}.  
We follow existing methods \cite{PADA} and select the first 25 classes images, in alphabetical order, as the target domain, and all 65 classes images from the other domain as source domain.

\subsection{Datasets \& Experimental Setup}

\begin{table*} 
\centering
\caption{Comparisons of Recognition Rates ($\%$) of Partial Domain Adaptation on Office31 Dataset (ResNet-50)} 
\label{Office31-10-res} 
\setlength{\tabcolsep}{3pt} % Default value: 6pt 
\renewcommand{\arraystretch}{1} % Default value: 1 
\begin{tabular}{|c|cccccc|c|} 
\hline 
Method & A31 $\rightarrow$ W10 & A31 $\rightarrow$ D10 & W31 $\rightarrow$ A10 & W31 $\rightarrow$ D10 & D31 $\rightarrow$ A10 & D31 $\rightarrow$ W10 & Average \\ \hline 
ResNet \cite{ResNet}& 75.59 & 83.44 & 84.97 & 98.09 & 83.92 & 96.27 & 87.05 \\ 
DAN \cite{long2015learning}& 59.32 & 61.78 & 67.64 & 90.45 & 74.95 & 73.90 & 71.34 \\ 
DANN \cite{DANN}& 73.56 & 81.53 & 86.12 & 98.73 & 82.78 & 96.27 & 86.50 \\ 
ADDA \cite{ADDA}& 75.67 & 83.41 & 84.25 & \underline{99.85} & 83.62 & 95.38 & 87.03 \\  
RTN \cite{long2016unsupervised}& 78.98 & 77.07 & 89.46 & 85.35 & 89.25 & 93.22 & 85.56 \\ 
IWAN \cite{IWAN}& 89.15 & 90.45 & 94.26 & 99.36 & \textbf{95.62} & 99.32 & 94.69 \\ 
SAN \cite{SAN}& 93.90 & 94.27 & 88.73 & 99.36 & 94.15 & 99.32 & 94.96 \\ %
PADA \cite{PADA}& 86.54 & 82.17 & \underline{95.41} & \textbf{100.0} & 92.69 & 99.32 & 92.69 \\ % 
DRCN \cite{li2020deep} & \underline{90.80} & \underline{94.30} & 94.80 & \textbf{100.0} & \underline{95.20} & \textbf{100.0} & \underline{95.90}\\ 
% ETN \cite{ETN}& \underline{94.52} & \underline{95.03} & 94.64 & \textbf{100.00} & \textbf{96.21} & \textbf{100.00} & \underline{96.73} \\ % 
Ours & \textbf{95.93} & \textbf{98.09} & \textbf{95.51} & \textbf{100.0} & 95.09 &\underline{99.66} & \textbf{97.38} \\ 
\hline 
\end{tabular} 
\end{table*}
\begin{table*} 
\centering
\caption{Comparisons of Recognition Rates ($\%$) of Partial Domain Adaptation on Office+Home Dataset (ResNet-50)} 
\label{OfficeHome} 
\setlength{\tabcolsep}{2pt} % Default value: 6pt 
\renewcommand{\arraystretch}{1} % Default value: 1 
\begin{tabular}{|c|cccccccccccc|c|} 
\hline 
Method & Ar $\rightarrow$ Cl & Ar $\rightarrow$ Pr & Ar $\rightarrow$ Rw & Cl $\rightarrow$ Ar & Cl $\rightarrow$ Pr & Cl $\rightarrow$ Rw & Pr $\rightarrow$ Ar & Pr $\rightarrow$ Cl & Pr $\rightarrow$ Rw & Rw $\rightarrow$ Ar & Rw $\rightarrow$ Cl & Rw $\rightarrow$ Pr & Average \\ \hline 
ResNet \cite{ResNet}& 46.33 & 67.51 & 75.87 & 59.14 & 59.94 & 62.73 & 58.22 & 41.79 & 74.88 & 67.40 & 48.18 & 74.17 & 61.35 \\ 
DAN \cite{long2015learning}& 43.76 & 67.90 & 77.47 & 63.73 & 58.99 & 67.59 & 56.84 & 37.07 & 76.37 & 69.15 & 44.30 & 77.48 & 61.72 \\
DANN \cite{DANN}& 45.23 & 68.79 & 79.21 & \underline{64.56} & 60.01 & 68.29 & 57.56 & 38.89 & 77.45 & 70.28 & 45.23 & 78.32 & 62.82 \\
ADDA \cite{ADDA}& 45.23 & 68.79 & 79.21 & \underline{64.56} & 60.01 & 68.29 & 57.56 & 38.89 & 77.45 & 70.28 & 45.23 & 78.32 & 62.82 \\
RTN \cite{long2016unsupervised}& 49.31 & 57.70 & 80.07 & 63.54 & 63.47 & 73.38 & 65.11 & 41.73 & 75.32 & 63.18 & 43.57 & 80.50 & 63.07 \\
IWAN \cite{IWAN}& 53.94 & 54.45 & 78.12 & 61.31 & 47.95 & 63.32 & 54.17 & \underline{52.02} & \underline{81.28} & \underline{76.46} & 56.75 & \underline{82.90} & 63.56 \\
SAN \cite{SAN}& 44.42 & 68.68 & 74.60 & \textbf{67.49} & \underline{64.99} & \textbf{77.80} & 59.78 & 44.72 & 80.07 & 72.18 & 50.21 & 78.66 & 65.30 \\
PADA \cite{PADA}& 51.95 & 67.00 & 78.74 & 52.16 & 53.78 & 59.03 & 52.61 & 43.22 & 78.79 & 73.73 & 56.60 & 77.09 & 62.06 \\
DRCN \cite{li2020deep} & \underline{54.00} & \underline{76.40} & \textbf{83.00} & 62.10 & 64.50 & 71.00 & \textbf{70.80} & 49.80 & 80.50 & \textbf{77.50} & \textbf{59.10} & 79.90 & \underline{69.00} \\ 
% ETN \cite{ETN}& \underline{59.24} & \underline{77.03} & 79.54 & 62.92 & \underline{65.73} & 75.01 & \underline{68.29} & \textbf{55.37} & \textbf{84.37} & \underline{75.72} & \underline{57.66} & \textbf{84.54} & \underline{70.45} \\
Ours &\textbf{60.30} & \textbf{80.17} & \underline{81.23} & \textbf{67.49} & \textbf{68.24} & \underline{76.04} & \underline{68.31} & \textbf{55.05} & \textbf{83.77} & 75.39 & \underline{58.93} & \textbf{83.14} & \textbf{71.51} \\
\hline 
\end{tabular}
\end{table*}

\begin{figure*}[!t]
\centering
\includegraphics[width=1\linewidth]{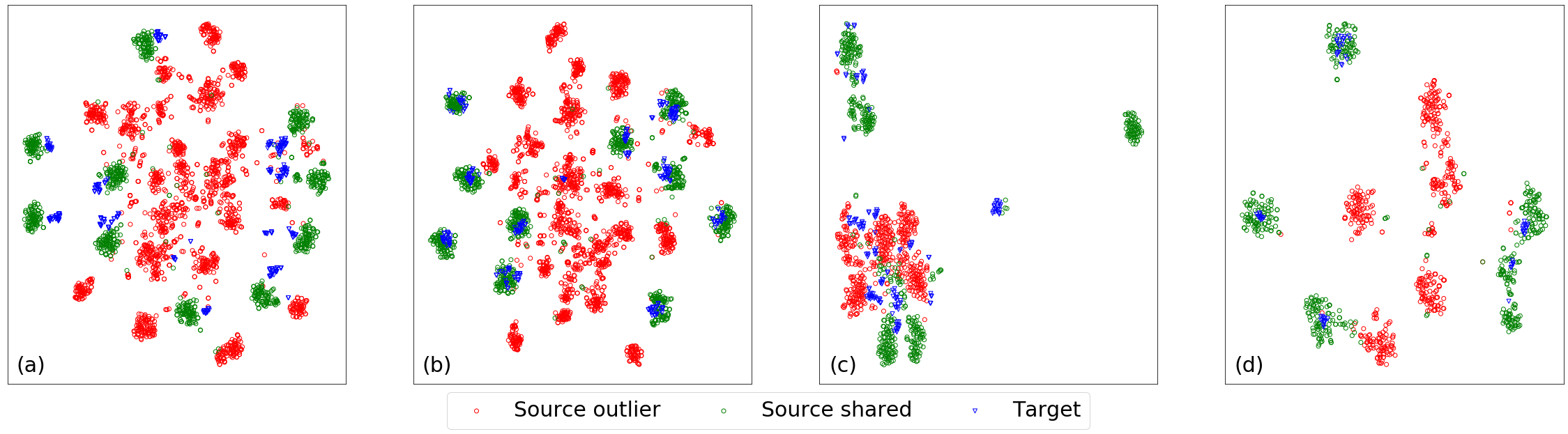}
\vspace{-6mm}
    \caption{(a) and (b) t-SNE visualization of original and embedding features from Amazon31 $\rightarrow$ Webcam10 task. (c) and (d) t-SNE visualization of original and embedding features from Amazon10 $\rightarrow$ DSLR5 task. } \vspace{-3mm}
    \label{fig:visual}
\end{figure*}

\noindent{\textbf{Comparisons}}: We compare our proposed model with several traditional domain adaptation models and partial domain adaptation methods: Deep Adaptation Network (DAN) \cite{long2015learning}, Reverse Gradient (RevGrad) \cite{ganin2015unsupervised}, Residual Transfer Network (RTN) \cite{long2016unsupervised}, Adversarial Discriminative Domain Adaptation (ADDA) \cite{ADDA}, Selective Adversarial Network (SAN) \cite{SAN}, Importance Weighted Adversarial Nets (IWAN) \cite{IWAN}, Partial Adversarial Domain Adaptation (PADA) \cite{PADA}, and Deep Residual Correction Network (DRCN) \cite{li2020deep}. We further compare with Convolutional Neural Network (AlexNet) \cite{krizhevsky2012imagenet} and Residual Network (ResNet) \cite{ResNet} as baselines. Specifically, DAN learns transferable features by matching different distributions optimally using multi-kernel MMD. RevGrad improves domain adaptation by making the source and target domains indistinguishable for a discriminative domain classifier via an adversarial training paradigm. RTN jointly learns transferable features and adapts different source and target classifiers via deep residual learning. ADDA combines discriminative modeling, untied weight sharing, and a GAN loss to yield much better results than RevGrad. SAN and IWAN  select or re-weighting outlier categories in source domain label space. PADA and DRCN are state-of-the-art partial domain adaptation models by exploring adversarial learning. PADA alleviates negative transfer through down-weighting the data of outlier source classes, DRCN explores residual block to promote the cross-domain feature representation learning and couples two domains by match shared classes feature distributions.
% while ETN proposes a progressive weighting scheme to quantify the transferability of source examples.

\noindent{\textbf{Implementation Details}}: Following the settings of IWAN \cite{IWAN}, we also adopt ImageNet pre-trained AlexNet \cite{krizhevsky2012imagenet} to obtain DeCAF$_6$ \cite{donahue2014decaf} features for Office\&Caltech-10 dataset, and source-finetuned ResNet-50 features for Office+Home dataset, with all images as 4096-dimension and 2048-dimension features, respectively. We also evaluate DeCAF$_6$ and ResNet-50 features on Office-31 dataset. Linear kernel is applied to DeCAF$_6$ features, while no kernel is applied to ResNet features. All methods are implemented with PyTorch and MATLAB. For parameter settings, we empirically set $\lambda = 0.1$, embedding features dimension $k = 100$, iteration number to 10 as default for all different tasks. The cross-domain graph $G$ in label refinement process is fully-connected, we choose cosine distance to build the weighted graph, and $\sigma=0.1$ on Office-31 and Office+Home dataset, while $\sigma=0.2$ on Office\&Caltech-10 tasks. We select $\alpha_p$ and $\alpha_c$ through 5-fold cross-validation \cite{liang2018aggregating} on the labeled source domain. %We will discuss the influence of different parameters in Section \ref{para}.

\subsection{Comparison Results \& Analysis}

Table \ref{OfficeCaltech10-5} and Table \ref{Office31-10} show the comparisons on partial domain adaptation tasks with DeCAF$_6$ features on Office\&Caltech-10 and Office-31 datasets, respectively. Results of tasks with ResNet-50 features on Office-31 and Office+Home dataset are shown in Table \ref{Office31-10-res} and Table \ref{OfficeHome}, respectively. The proposed DCDF framework achieves the best average classification accuracy on all three tasks, and outperforms state-of-the-art partial domain adaptation methods on most cases, which prove the effectiveness of DCDF on different datasets across different features. On Office\&Caltech-10 $\rightarrow$ 5 tasks, it is noteworthy that our proposed model achieves $100\%$ accuracy in two cases: W10 $\rightarrow$ D5 and D10 $\rightarrow$ W5, and performs $5\%$ higher than IWAN, and $4\%$ higher than ADDA in case A10 $\rightarrow$ D5. Moreover, on the tasks with more categories and samples in Office-31 dataset with DeCAF$_6$ features, our proposed method outperforms almost all the cases and achieves more than $3\%$ improvements on average accuracy, and more than $5\%$ higher in several cases, e.g., A31 $\rightarrow$ W10, than partial domain adaptation methods IWAN. For ResNet-50 features, due to the contribution of pre-trained and finetuned ResNet model, the baseline results of Office-31 and Office+Home tasks are much higher than AlexNet based DeCAF$_6$ features, while our proposed method DCDF still performs better than state-of-the-art partial domain adaptation model in most cases.

\begin{figure*}[!t]
\centering
\includegraphics[width=1\linewidth, scale=0.8]{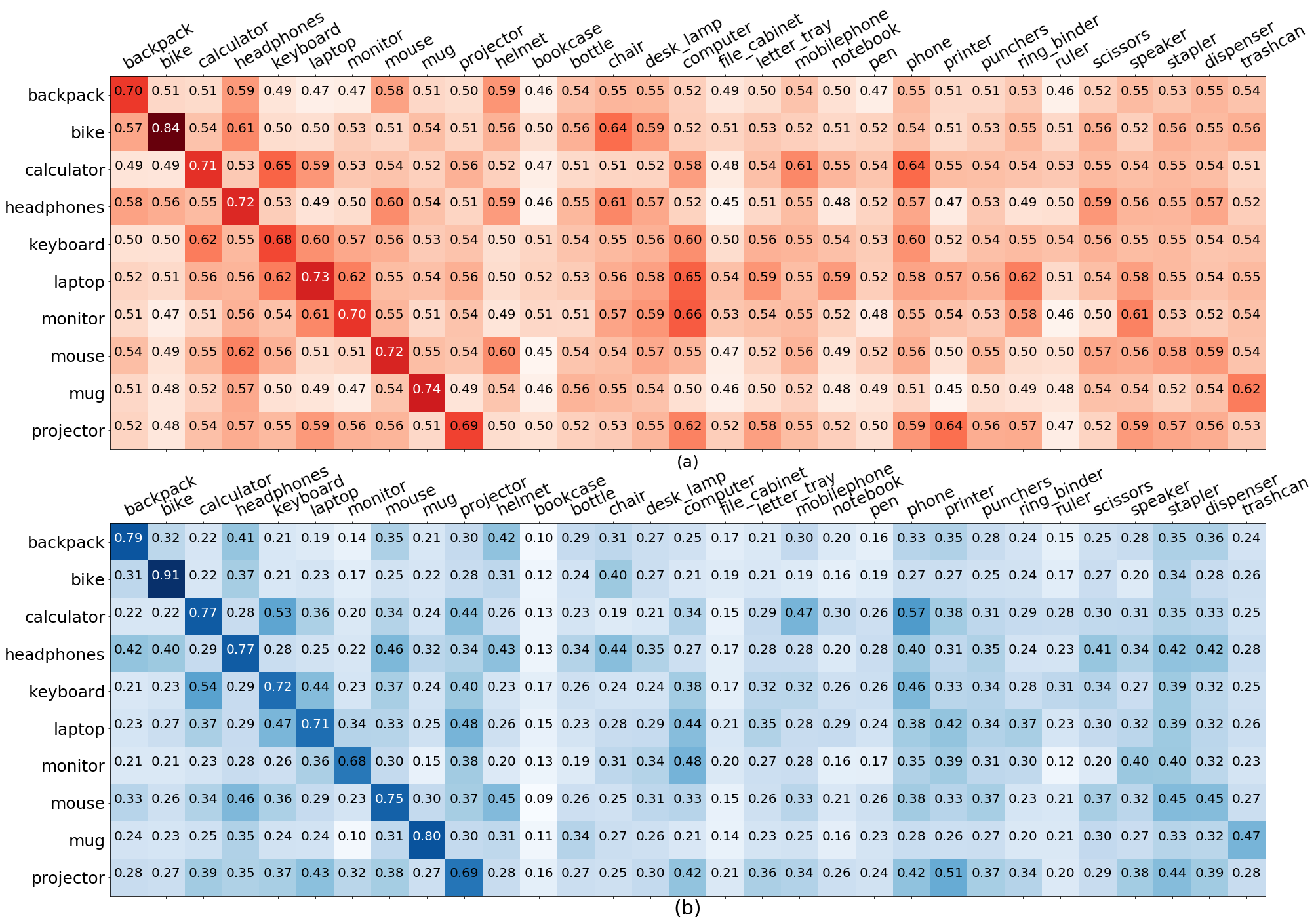}
    \caption{Cosine similarity matrix on Office31 dataset A31 $\rightarrow$ W10 case. (a) Cosine similarity matrix between source and target domain original features. (b) Cosine similarity matrix of source and target domain embedding features after DCDF adaptation}
    \label{fig:similarity}
\end{figure*}

\begin{figure*}[!t]
\centering
\includegraphics[width=1\linewidth, scale=0.8]{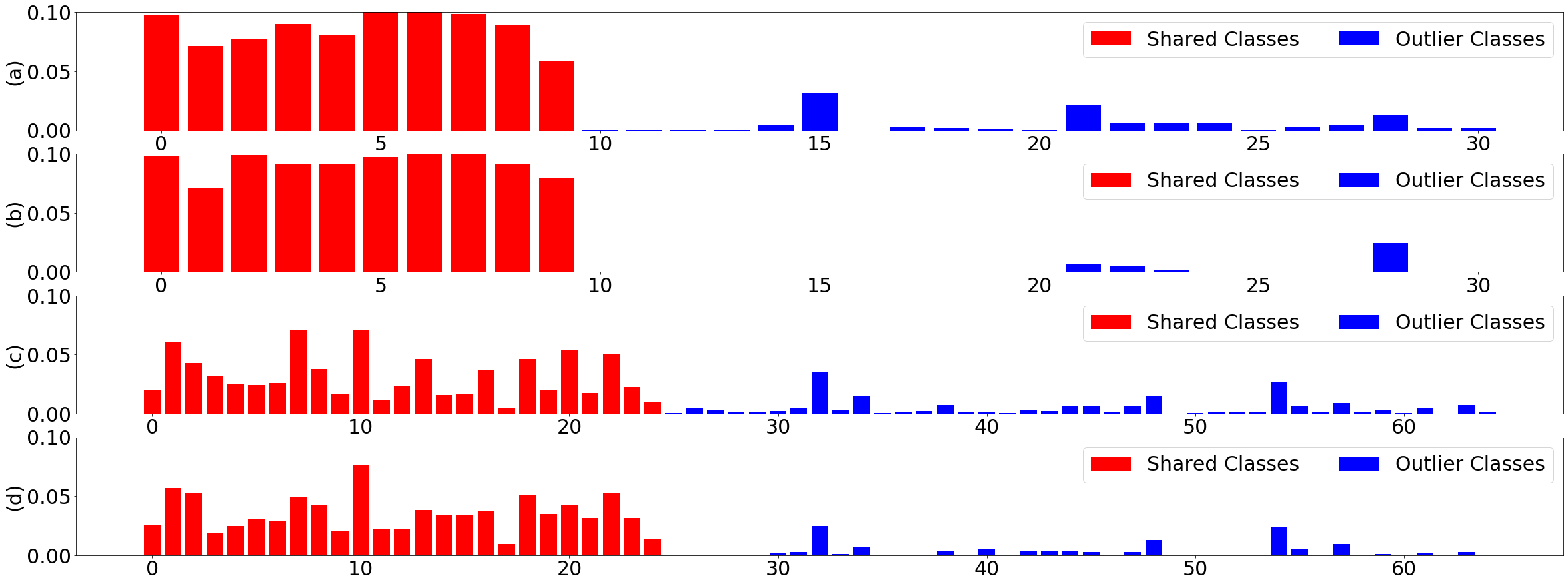}
    \caption{(a) and (b) Predicted class weights with original and embedding features on Office dataset Amazon 31 $\rightarrow$ Webcam 10 task. (c) Predicted class weights of original and embedding features on Office+Home dataset Art 65 $\rightarrow$ Product 25 task. }\vspace{-3mm}
    \label{fig:weights}
\end{figure*}

First, AlexNet and ResNet are pre-trained on ImageNet dataset and fine-tuned on source domain only, which makes it perform comparably with domain adaptation approaches only when source and target domain shift is marginal, e.g., W10 $\rightarrow$ D5. However, when source and target domain shift is large, source-only trained AlexNet and ResNet cannot handle it anymore, such as A10 $\rightarrow$ W5.

\begin{figure*}[!t]
\centering
\includegraphics[width=1\linewidth]{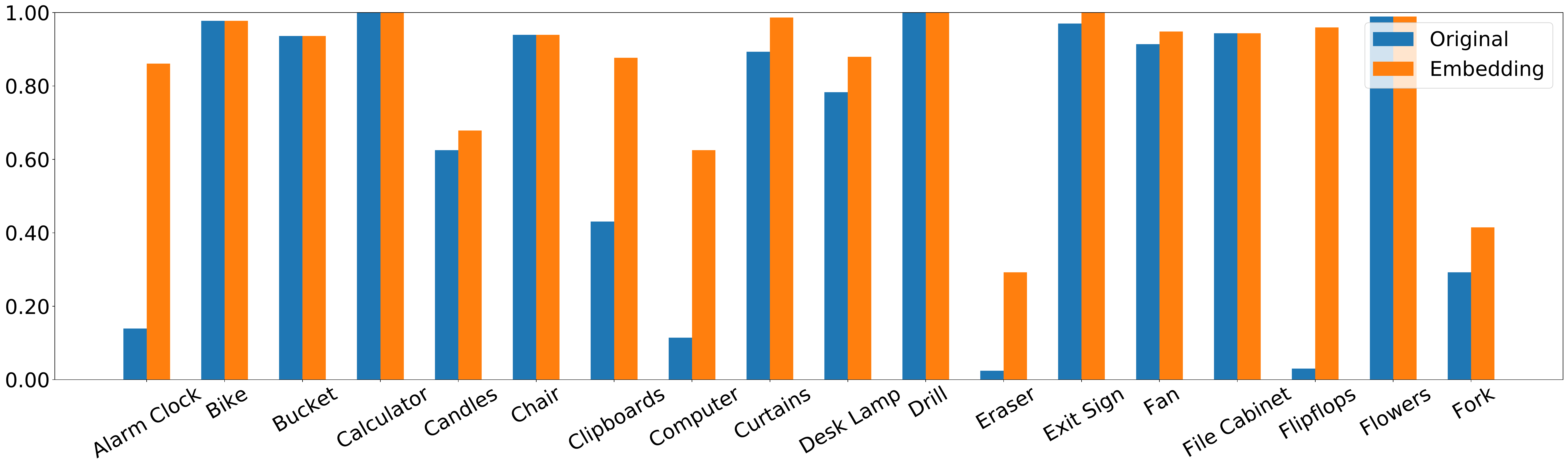}
    \caption{Classification accuracy results of each category on Office+Home dataset case Art65 $\rightarrow$ Product25 with original features (blue color) and after embedding features (orange color) through DCDF adaptation.}
    \label{fig:accs}  \vspace{-6mm}
\end{figure*}

Second, DAN is MMD-based network, which seeks to eliminate the disparity of source and target feature distribution and spread target samples to all source label space. Since the influence of negative transfer, DAN even performs worse than AlexNet on many tasks. RevGrad implements adversarial network and domain classifier to enhance the classification task. Similar to DAN, RevGrad only seeks to minimize marginal distribution difference between source and target domain without considering conditional distribution disparity, which also generates poor performance on partial domain adaptation tasks. RTN introduces entropy minimization criterion and residual block to preserve the target domain data structure, which would minimize the impact of outlier source domain categories to some degree. However, the results in table \ref{OfficeCaltech10-5} reveal that RTN cannot avoid negative transfer effectively.

\begin{figure*}[t]
\centering
\includegraphics[width=1\linewidth]{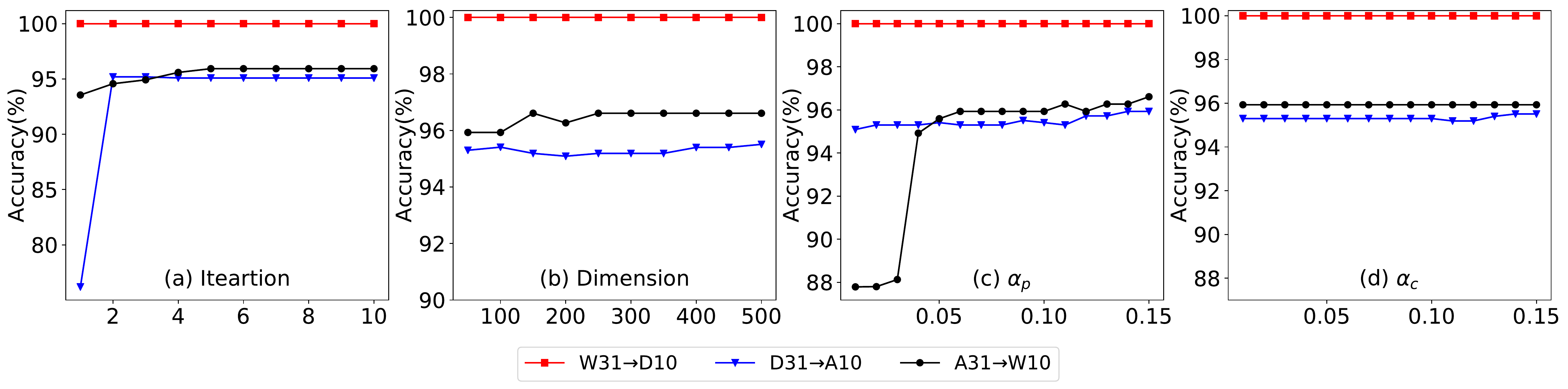}
    \caption{(a) Iterative results of our model. (b) Dimensionality analysis of $\mathbf{A}$. (c) Parameter $\alpha_p$ influence. (d) Parameter $\alpha_c$ influence.}
    \label{fig:parameters}  \vspace{-3mm}
\end{figure*}

Third, ADDA and IWAN have similar idea and network structure, and ADDA can be treated as non-weighted special case of IWAN. IWAN proposes a strategy of two classifiers in adversarial networks to identify if the source samples are from shared label space or outliers, which will mitigate domain shift and benefit partial domain adaptation tasks. However, IWAN obtains impressive results on small dataset tasks, e.g. Office10 $\rightarrow$ 5, rather than large-scale datasets like Office-31 $\rightarrow$ 10, which proves that only re-weighting domains to target samples cannot alleviate the influence of outlier source and domain shift on large dataset. ADDA is IWAN without domain weighting strategy, which also spreads target domain samples to all source label space, and suffers from negative transfer of outlier source.

Finally, SAN, PADA, and DRCN are most recent partial domain adaptation algorithms, so they perform better than baselines and other domain adaptation methods through reweighting shared and outlier classes to alleviate negative transfer. However, SAN implements a lot of classifiers in network, which means a large number of parameters involved. PADA applies adversarial networks and minmax optimization to re-weight the influence of source domain samples but only re-weights the source class level contribution. DRCN enhances the cross-domain adaptation and boosts the feature representation capability by plugging a residual block into the networks and weakens the irrelevant classes misleading with the help of a weighted class-wise alignment loss, but the experiments results illustrate that DRCN still cannot compete with our model on most tasks.
% while ETN proposes a progressive example transferability quantification method to reweight each source samples contribution to partial domain adaptation. 
Our proposed DCDF combines the class-wise and samples-level reweighting idea, as well as graph-based cross-domain structural knowledge transfer mechanism, which obtains the best performance on most cases. Moreover, SAN, IWAN, are all GAN-based frameworks, while our proposed model DCDF seeks a domain-invariant subspace over deep features, which makes our training and optimization process efficient and effective. The results in Table \ref{Office31-10} and Table \ref{Office31-10-res} verify the effectiveness of our model over different deep features with a further knowledge transfer stage.

\subsection{Empirical Analysis}\label{para}

In this part, we discuss some properties of our proposed model and results evaluation.

First of all, we visualize the projected embedding features distribution of source and target samples from Office31 dataset on Amazon31 $\rightarrow$ Webcam10, and Office\&Caltech-10 dataset on Amazon10 $\rightarrow$ DSLR5. As shown in Fig. \ref{fig:visual}, green and red circles denote shared and outlier source domain samples, respectively, while blue inverted triangles represent target domain samples. Figs. \ref{fig:visual} (a)(c) show the data distribution before domain adaptation through DCDF, while Figs. \ref{fig:visual} (b)(d) visualize the features distribution after domain adaptation through our proposed model. From the results, we observe that the target domain samples, blue inverted triangles, are well aligned to the green circles class centers rather than spreading to all source domain label space. This indicates the effectiveness of our proposed method in aligning target data to relevant source classes and eliminating negative transfer caused by outliers.

Secondly, we take Amazon31 $\rightarrow$ Webcam10 from Office31 dataset to calculate similarity matrix across original and embedding features. In Fig. \ref{fig:similarity}, we show the average cosine similarity of samples from the same category. We notice that after domain adaptation, the similarities between source and target domain embedding features from the same classes are getting larger, while the similarities between irrelevant classes are getting significantly smaller, which proves that our DCDF simultaneously pulls the samples from the same classes closer, and pushes irrelevant classes further away. Actually, deep network parameters pre-trained on large-scale a dataset already have a good generalization ability, which is also a hot topic to learn good representation by designing deep architectures. For domain adaptation, we found the knowledge transfer on the top layer is the key issue. That is why our two-step strategy could still achieve better performance than end-to-end learning models.

Thirdly, Fig. \ref{fig:weights} illustrates the class weights learned by our proposed method on case Amazon31 $\rightarrow$ Webcam10 and case Art65 $\rightarrow$ Product25. It is noticeable that with the same label propagation method and parameter settings, for the original features versus adapted embedding features, the outlier classes weights is getting smaller or even removed totally. This demonstrates the effectiveness of cross-domain weighted graph label propagation to identify source and target domain shared classes from outliers. Moreover, we further list the similarity comparison of each category over Art65 $\rightarrow$ Product25 in Fig. \ref{fig:accs}, where we notice that our model significantly improves the accuracy of some categories over the original features. This indicates the effectiveness in mitigating the domain mismatch and enhancing the recognition accuracy.

Finally, we present the iterative performance, parameters sensitivity and dimensionality influence of $\mathbf{A}$ on Office-31 $\rightarrow$ 10 tasks in Fig. \ref{fig:parameters}. In our experiments, we notice that most cases reach the final position within 5 iterations, which verifies our approache converges very well (Fig. \ref {fig:parameters}-a). For dimensionality influence (Fig. \ref {fig:parameters}-b), we observe that for cases like W31 $\rightarrow$ D10 and D31 $\rightarrow$ A10, the embedding features dimension does not influence the performance too much. In cases like A31 $\rightarrow$ W10, our proposed method works well when the dimension is low. And with the increase of the dimension, the performance even increase a little bit. We assume that the embedding features would contribute more information with the dimension increasing. Parameters $\alpha_p$ and $\alpha_c$ balance the contribution of different terms (Fig. \ref{fig:parameters}-c \& d). For some cases like W31 $\rightarrow$ D10, where the prediction performs favorably thus $\alpha_p$ would not affect the results very much. However, for those cases where predicted pseudo-labels are less accurate, e.g., A31 $\rightarrow$ W10, larger $\alpha_p$ would involve more probabilistic soft labels and avoid misleading optimization thoroughly. For $\alpha_c$, we notice it is not sensitive across different tasks. 

\section{Conclusion}

We proposed a novel discriminative cross-domain feature learning framework (DCDF) for partial domain adaptation, where external source domain covers more classes than the unlabeled target domain. Specifically, a selective domain-wise adaptation, weighted class-wise alignment, and the discriminative domain-invariant center loss are proposed to align unlabeled target data with source domain shared class centers, while a weighted cross-domain graph would capture the intrinsic structure within source and target to propagate source labels effectively to target samples. The experimental results on several cross-domain benchmarks proved the effectiveness of our proposed model in partial domain adaptation tasks over state-of-the-art methods.

\bibliographystyle{IEEEtran}
\bibliography{DCDF}

\end{document}